\documentclass[sigconf]{acmart}

\copyrightyear{2026}
\acmYear{2026}
\setcopyright{cc}
\setcctype{by}
\acmConference[WWW '26]{Proceedings of the ACM Web Conference 2026}{April 13--17, 2026}{Dubai, United Arab Emirates.}
\acmBooktitle{Proceedings of the ACM Web Conference 2026 (WWW '26), April 13--17, 2026, Dubai, United Arab Emirates}
\acmISBN{979-8-4007-2307-0/2026/04}
\acmDOI{10.1145/3774904.3792257}

\usepackage{threeparttable}
\usepackage{listings}
\usepackage{xcolor}
\usepackage{float}
\usepackage{bbding}   
\begin{document}

\title{Curiosity Driven Knowledge Retrieval for Mobile Agents}

\author{Sijia Li}
\authornote{Both authors contributed equally to this research.}
\email{sijialiyabing@gmail.com}
\orcid{0009-0005-0318-4276}
\affiliation{%
  \institution{Shanghai University of Engineering Science}
  \city{Shanghai}
  \country{China}
}

\author{Xiaoyu Tan}
\orcid{0000-0003-3555-7143}
\authornotemark[1]
\email{txywilliam1993@outlook.com}
\affiliation{%
  \institution{National University of Singapore}
  \country{Singapore}
}

\author{Shahir Ali}
\email{shahir@droidrun.ai}
\orcid{0009-0007-3651-5654}
\affiliation{%
  \institution{Droidrun}
  \city{Osnabrück}
  \country{Germany}
}

\author{Niels Schmidt}
\email{niels@droidrun.ai}
\orcid{0009-0003-4950-6555}
\affiliation{%
  \institution{Droidrun}
  \city{Osnabrück}
  \country{Germany}
}

\author{Gengchen Ma}
\email{m320123320@sues.edu.cn}
\orcid{0009-0002-3841-9889}
\affiliation{%
  \institution{Shanghai University of Engineering Science}
  \city{Shanghai}
  \country{China}
}

\author{Xihe Qiu}
\authornote{Corresponding author.}
\orcid{0000-0003-4024-925X}
\email{qiuxihe1993@gmail.com}
\affiliation{%
  \institution{Shanghai University of Engineering Science	}
  \city{Shanghai}
  \country{China}
}

\renewcommand{\shortauthors}{Li et al.}

\begin{abstract}
Mobile agents have made progress toward reliable smartphone automation, yet performance in complex applications remains limited by incomplete knowledge and weak generalization to unseen environments. We introduce a curiosity driven knowledge retrieval framework that formalizes uncertainty during execution as a curiosity score. When this score exceeds a threshold, the system retrieves external information from documentation, code repositories, and historical trajectories. Retrieved content is organized into structured \textit{AppCards}, which encode functional semantics, parameter conventions, interface mappings, and interaction patterns. During execution, an enhanced agent selectively integrates relevant \textit{AppCards} into its reasoning process, thereby compensating for knowledge blind spots and improving planning reliability. Evaluation on the AndroidWorld benchmark shows consistent improvements across backbones, with an average gain of six percentage points and a new state of the art success rate of 88.8\% when combined with GPT-5. Analysis indicates that AppCards are particularly effective for multi step and cross application tasks, while improvements depend on the backbone model. Case studies further confirm that AppCards reduce ambiguity, shorten exploration, and support stable execution trajectories. Task trajectories are publicly available at \url{https://lisalsj.github.io/Droidrun-appcard/}.
\end{abstract}



\maketitle

\begin{figure*}
\centering
\includegraphics[width=1\linewidth]{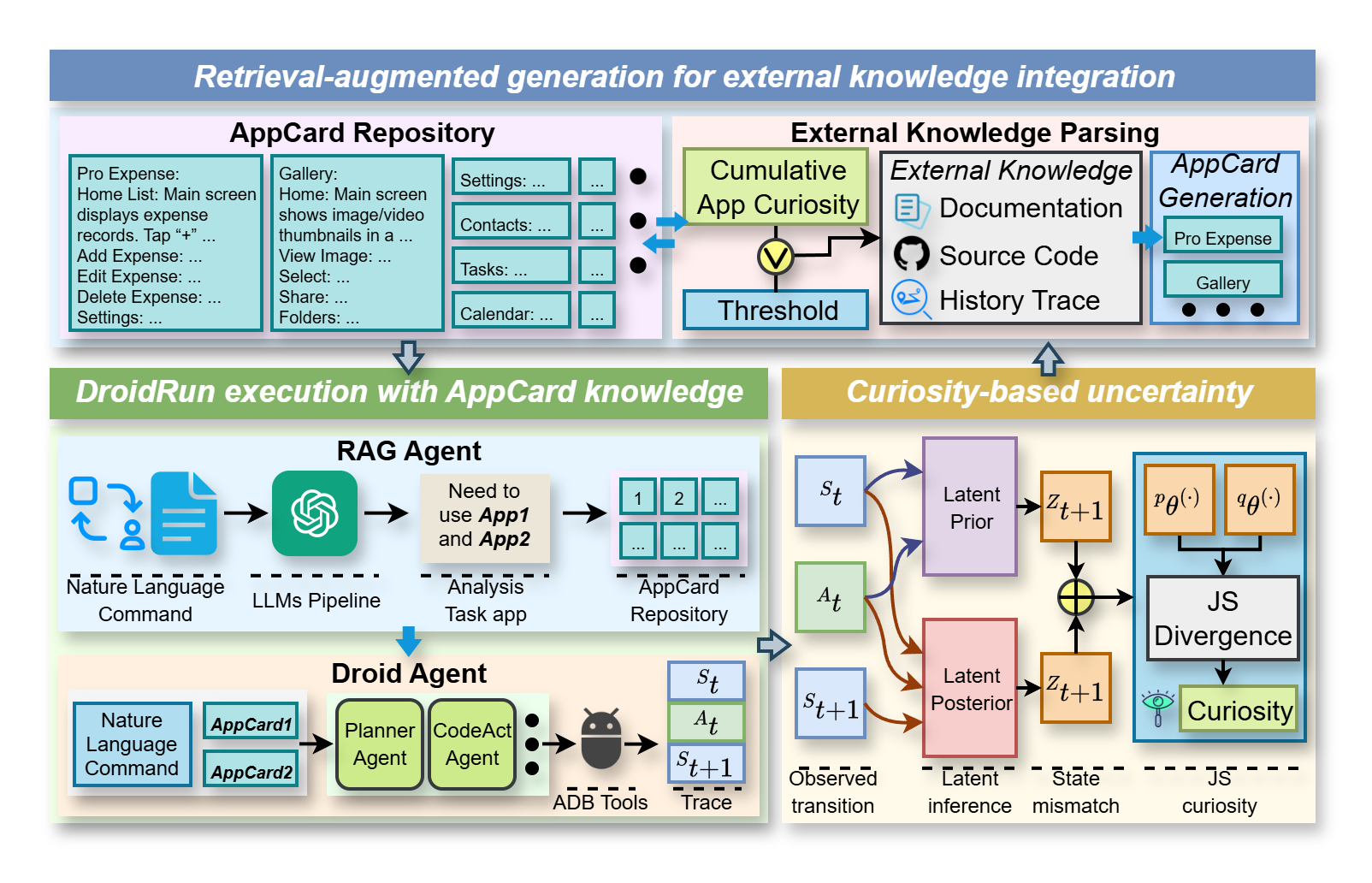}
\caption{Overall framework of the curiosity driven knowledge retrieval system for mobile agents. Task execution is guided by AppCards. Uncertainty estimation produces curiosity signals that activate external retrieval. Retrieved knowledge is consolidated to update AppCards and the updated AppCards are reintegrated into the execution pipeline.}
\label{main_framework}
\end{figure*}

\section{Introduction}
\label{sec:intro}
Mobile agents aim to deliver convenient and intelligent automation on smartphones. Given a natural language instruction, an agent infers user intent, constructs an execution plan, and orchestrates actions across installed applications to accomplish the specified goal. Advances in multimodal large language models \cite{zhang2024mm} have markedly improved interaction competence, task understanding, and coverage of application ecosystems, indicating strong potential for intelligent interaction and everyday automation \cite{masterman2024landscape}. Despite this progress, practical deployments reveal important limitations. When interacting with complex applications or with environments that have not been encountered before, agents often lack application specific functional knowledge. This knowledge incompleteness produces planning errors, incorrect invocation of applications or tools, and reduced task success, and it limits generalization to new tasks and applications.


Existing approaches to strengthening task execution in mobile agents fall into two broad lines \cite{tang2025survey}. The first line emphasizes application oriented end to end training in which large scale interaction data from one or a small number of target applications is used for supervised fine tuning SFT \cite{anisuzzaman2025fine} or reinforcement fine tuning RFT \cite{zhai2024fine}. These methods yield substantial gains within the environments used for training yet they transfer poorly to previously unseen applications. Typical failure modes include overfitting to specific interface layouts and action vocabularies, mismatch between learned policies and new widget taxonomies, and brittle behavior under distribution shift. A complementary line focuses on framework level improvements that aim to build stable and general agent architectures for planning and reasoning \cite{guo2024large, hao2025reflective}. Although these designs improve the handling of complex interactive tasks and support longer execution horizons, they usually encode only shallow or implicit knowledge of application functionality. As a result the agent often issues erroneous API or function calls, including incorrect entry points and misparameterized arguments, which propagate through the interaction sequence and ultimately limit overall task completion.

In this work a curiosity driven knowledge retrieval framework is introduced to strengthen task execution of mobile agents in complex application scenarios. A curiosity score is formalized to quantify the agent’s insufficient understanding during application interactions as uncertainty over functional knowledge. When this score for a given application exceeds a predefined threshold \cite{he2024curiosity}, the system initiates external knowledge retrieval \cite{huang2022unified}. The curiosity gate serves as a resource allocation mechanism that bounds retrieval frequency, preserves context window budget, and concentrates attention on states with the largest epistemic deficit. The retrieval process extracts salient information from application documentation and source repositories and organizes the material into structured \textit{AppCards}. During execution the agent analyzes the task, identifies the relevant applications, and invokes the corresponding \textit{AppCards} so that application knowledge is incorporated into reasoning and decision making. Because \textit{AppCards} are modular and version aware, only the fragments required by the current step are injected, which reduces spurious associations and improves stability under user interface changes. This pipeline yields a finer grained representation of application functionality at the framework level and improves planning reliability, automation efficiency, and task success in challenging scenarios.

Specifically, our contributions can be summarized as follows:  
\begin{enumerate}
    \item A curiosity driven knowledge augmentation framework is presented, which quantifies insufficient understanding as a curiosity score and uses this signal as a principled trigger for external knowledge retrieval.  
    \item A structured representation termed \textit{AppCards} is developed to consolidate knowledge from documentation and repositories, enabling efficient integration and enhancing fine grained understanding of application functionalities.  
    \item Comprehensive evaluation on the AndroidWorld benchmark \cite{rawles2024androidworld} with the DroidRun agent demonstrates consistent gains in task success and robustness, with further validation across unseen applications and multiple backbone models.  
\end{enumerate}

\section{Related Work}
\subsection{Mobile Agents and GUI Automation}
In the study of GUI automation and mobile agents, early approaches primarily relied on scripting and demonstration-driven strategies. RecurBot \cite{intharah2018recurbot} enabled automatic completion of repetitive operations by identifying loop patterns within demonstrations, while testing frameworks such as Appium and SilkTest \cite{appium_github, silktest} provided script-based cross-platform interaction capabilities. With the advancement of multimodal models, research has increasingly shifted toward unified frameworks that integrate visual and linguistic information. Aria-UI \cite{yang2024aria} achieved vision-only grounding through large-scale synthetic data and context modeling, and SeeClick \cite{cheng2024seeclick} introduced GUI pre-training together with the ScreenSpot benchmark. UI-TARS \cite{qin2025ui} and AndroidGen \cite{lai2025androidgen} explored instruction-to-element mapping in diverse application scenarios using multimodal models, whereas GUI-Explorer \cite{xie2025gui} improved cross-task generalization by combining screenshots with accessibility trees. Building on these efforts, more recent studies emphasize the design of agent frameworks; for example, MobileUse \cite{li2025mobileuse} employs a hierarchical reflection mechanism to enhance stability and adaptability in task execution. Collectively, these works have advanced mobile agents toward more reliable understanding of natural language instructions and robust interaction with user interfaces across applications and platforms.

\subsection{Knowledge Augmentation and Reasoning Frameworks}
In the context of knowledge augmentation for agents, LLM-Explorer introduces a knowledge maintenance–centric framework that leverages large language models to construct and update abstract interaction graphs, thereby achieving high interface coverage while substantially reducing inference calls and associated costs \cite{zhao2025llm}. However, its effectiveness remains limited in scenarios requiring cross-application transitions and state sharing. Recent work has emphasized large-scale interaction data pretraining to enhance the generality and transferability of agents. AutoGLM develops a deployable foundation agent system across representative GUI domains such as browsers and mobile applications, integrating perception, planning, and execution modules to strengthen cross-application capabilities \cite{liu2024autoglm}, while GUI-Owl-32B/7B serves as a foundation model that demonstrates robust end-to-end performance and portability across both desktop and mobile environments \cite{ye2025mobile}. Moreover, although SymAgent is not designed specifically for mobile agents, its neural symbolic self-learning framework treats knowledge graphs as dynamic reasoning environments. By combining a planner and an executor architecture with external retrieval augmentation, it improves complex task decomposition and execution stability \cite{liu2025symagent}. This approach provides useful insights into knowledge augmentation and reasoning for mobile agents in scenarios that involve complex reasoning and long-horizon tasks.

\subsection{Curiosity Driven Knowledge Exploration}
Curiosity in reinforcement learning is commonly viewed as an intrinsic mechanism that drives agents to explore uncertain states \cite{schmidhuber1991possibility}. A representative approach is the predictive model of Pathak et al., where large prediction errors indicate poorly captured dynamics and thus yield intrinsic rewards that encourage exploration \cite{pathak2017curiosity}. Recent work extends this idea to large language models, using RL-style formulations to acquire missing knowledge during reasoning \cite{yao2023react}. For example, WorldLLM combines Bayesian inference with curiosity-driven exploration to iteratively generate and validate interpretable world hypotheses \cite{levy2025worldllm}, while CDE defines curiosity rewards from perplexity-based actor signals and uncertainty estimated from multi-critic variance to improve RLVR training \cite{dai2025cde}. More broadly, uncertainty-guided mechanisms have been used to steer selective decision-making under distribution shift \cite{fang2024uncertainty}. Inspired by this view, we use curiosity as a lightweight trigger for targeted knowledge acquisition: when the agent is uncertain in a previously unseen application, it initiates external retrieval and constructs an AppCard to reduce future uncertainty with minimal manual effort.

\begin{figure*}[h]
\centering
\includegraphics[width=1\linewidth]{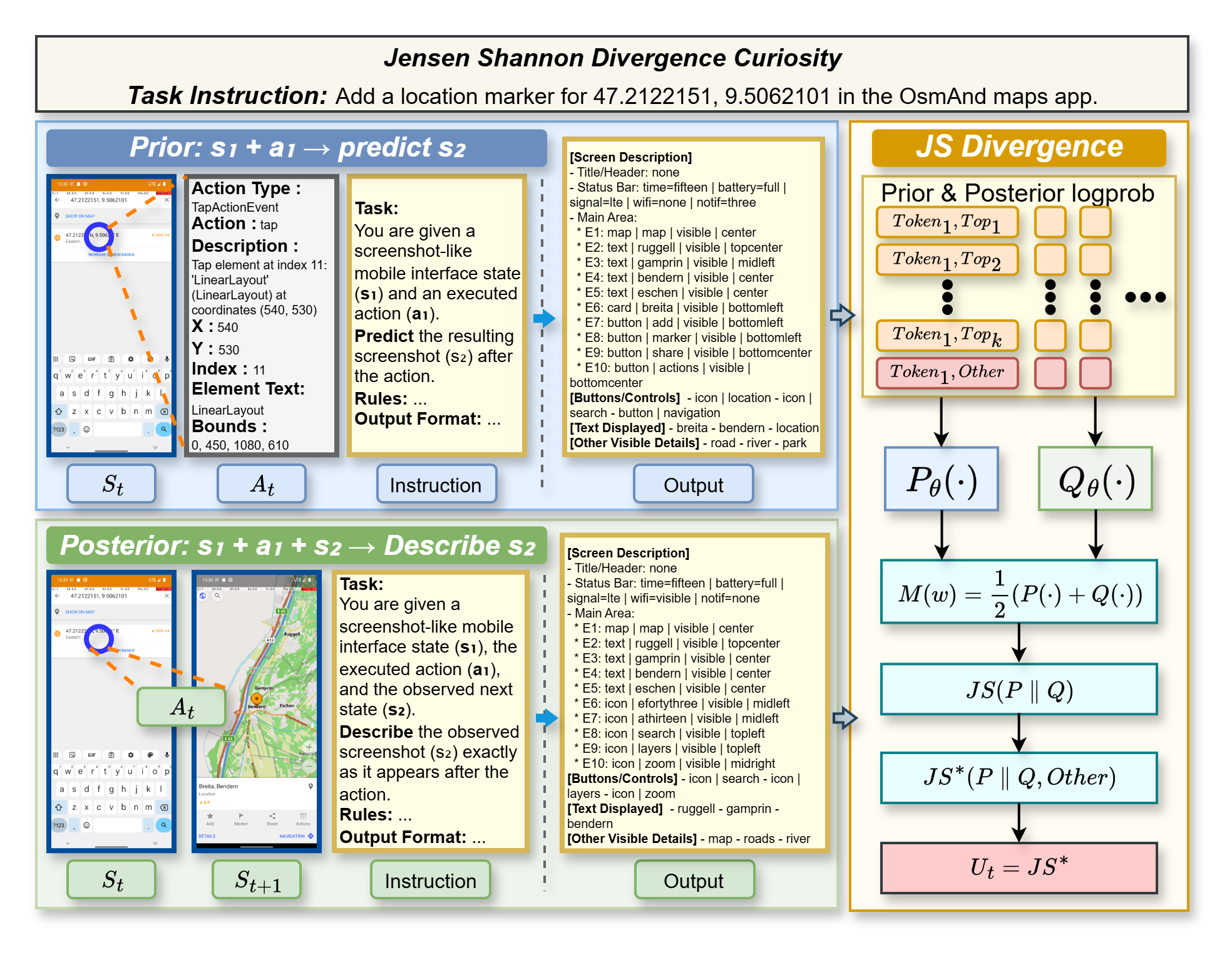}
\caption{JS-divergence based curiosity estimation. The agent predicts the next interface state from the current state and action as a prior distribution, then observes the next state as a posterior distribution. The divergence between these distributions is measured with a tail adjusted Jensen Shannon divergence, yielding an information gain signal quantifying curiosity.}
\label{curiosity}
\end{figure*}

\section{Methods}
\label{sec:methods}

\subsection{Problem Formulation}

We formulate mobile-agent task execution as a variant of the Markov Decision Process (MDP) \cite{bellman1957markovian}, represented by the five-tuple
\begin{equation}
    \mathcal{M} = \{ \mathcal{S}, \mathcal{A}, U, \tau, \gamma \}.
\end{equation}

Here, $\mathcal{S}$ denotes the state space, corresponding to the mobile interface observed by the agent. $\mathcal{A}$ represents the action space, including operations such as \textit{tap\_by\_index}, \textit{tap\_by\_text}, \textit{input\_text}, and \textit{swipe}. Different from the classical MDP, we introduce $U$ as a quantitative measure of the agent’s epistemic uncertainty about application functionalities, serving as the intrinsic signal for curiosity-driven exploration. $\tau$ denotes the exploration threshold: once accumulated uncertainty exceeds this value, the agent triggers external knowledge acquisition to refine its comprehension. Finally, $\gamma \in (0,1]$ is the discount factor that balances immediate and long-term rewards.

This formulation highlights the dual role of uncertainty. It reflects the agent’s cognitive limitations and governs the initiation of external knowledge retrieval, thereby enhancing robustness and generalization (Figure~\ref{main_framework}).


\subsection{Curiosity-Driven Information Gain Estimation}

When vision-language models act as agents to accomplish tasks, it is crucial to quantify their insufficiency in understanding the application environment in order to trigger external knowledge retrieval. Inspired by the concept of \textbf{Latent Bayesian Surprise (LBS)} \cite{mazzaglia2022curiosity}, we treat the divergence between the prior prediction and the posterior observation as the principal signal for measuring uncertainty. This divergence reflects the information gain acquired during interactions and can be interpreted as a curiosity signal that prompts the agent to proactively initiate external knowledge retrieval when facing knowledge gaps. This formulation is also aligned with recent advances CDE \cite{dai2025cde}, which demonstrate that explicitly modeling uncertainty as an intrinsic signal can effectively guide exploration and improve reasoning performance.

\subsubsection{Latent State Distribution Modeling}

Given the current interface state $s_t$ and the executed action $a_t$, the model generates a prior predictive distribution $P$ for the description of the next interface. After observing the actual subsequent state $s_{t+1}$, the model produces a posterior distribution $Q$ for the corresponding description. At each decoding step $t$, only the Top $K$ candidate tokens are retained, and the remaining probability mass is aggregated into a residual bucket denoted as OTHER. Formally,

\begin{equation}
p_{t,i} = \exp(\ell_{t,i}), \quad 
q_{t,i} = \exp(\hat{\ell}_{t,i}), \quad i=1,\dots,K
\end{equation}

where $\ell_{t,i}$ denotes the prior log probability of the $i$th token at decoding step $t$, and $\hat{\ell}_{t,i}$ denotes the posterior log probability of the $i$th token at the same step after incorporating $s_{t+1}$. The residual probabilities at each decoding step are defined as

\begin{equation}
p_{t,\mathrm{other}} = 1-\sum_{i=1}^K p_{t,i}, \quad 
q_{t,\mathrm{other}} = 1-\sum_{i=1}^K q_{t,i}
\end{equation}

Each output token distribution is then normalized so that the $K+1$ dimensional probability vector consisting of the Top $K$ tokens and the residual probability sums to one. By aggregating across all decoding positions, we define a union set of length $T$. For any token $w$, the probabilities are accumulated and normalized to obtain the final prior distribution and posterior distribution as

\begin{equation}
P(w)=\frac{{\sum_{t=1}^T p_t(w)}}{T}, \quad 
Q(w)=\frac{{\sum_{t=1}^T q_t(w)}}{T}
\end{equation}

\begin{figure}[h]
    \centering
    \includegraphics[width=1\linewidth]{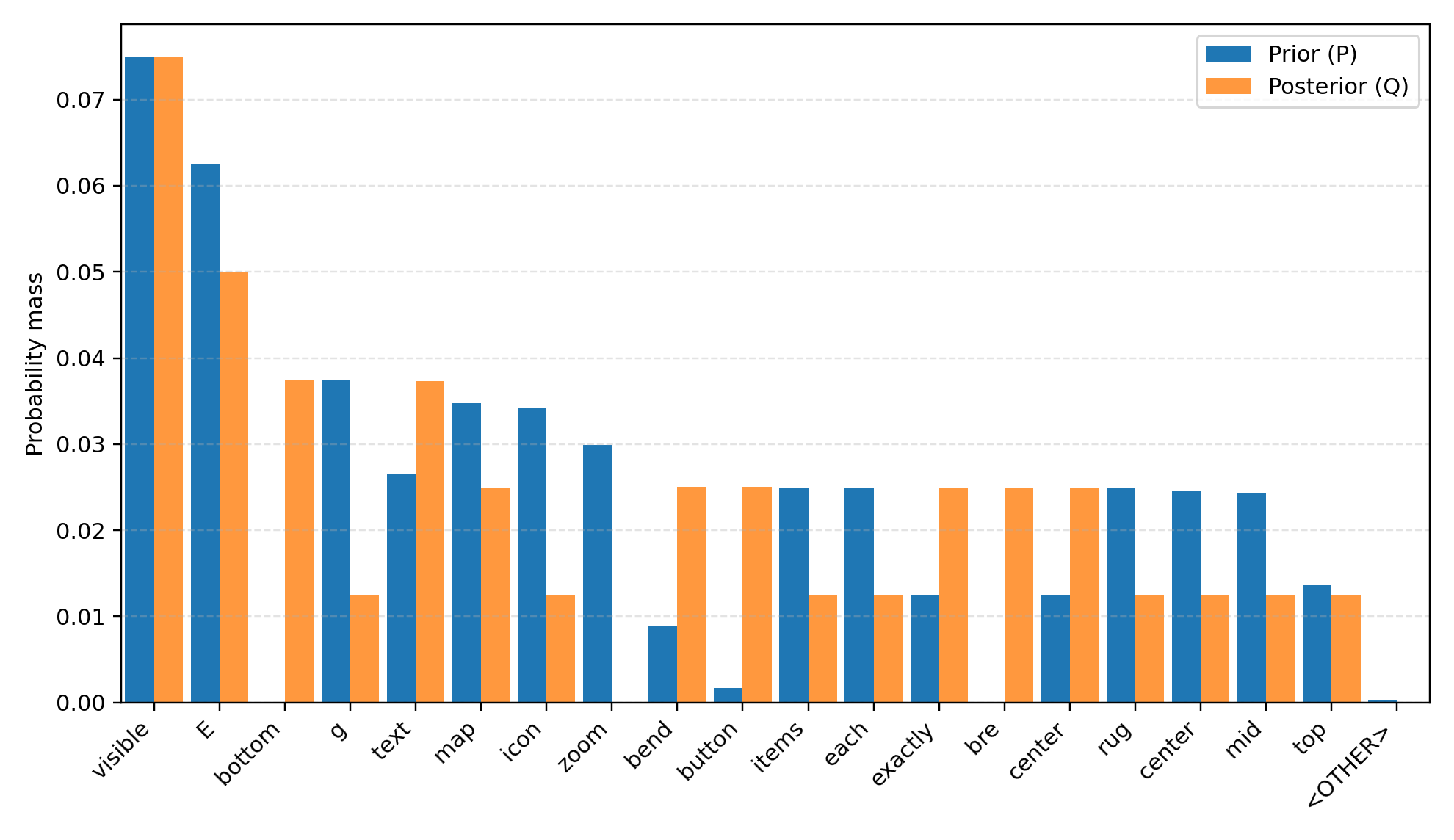}
    \caption{An example of aggregated global semantic distributions with Top-19 tokens and the residual OTHER category}
    \label{global_semantic}
\end{figure}

This procedure yields global prior and posterior distributions that capture token-level semantics across decoding steps. Figure~\ref{global_semantic} provides an illustrative example of such aggregated distributions, offering an intuitive view of the differences between $P$ and $Q$.

\subsubsection{Information Gain Estimation}

Within a Bayesian perspective, the information acquired by the agent during the interaction $\,s_t,a_t \rightarrow s_{t+1}\,$ is measured by the divergence between the prior and the posterior distributions
\begin{equation}
I(z_{t+1};\, s_{t+1}\mid s_t,a_t)\;\approx\; D_{\mathrm{KL}}(P\parallel Q)
\end{equation}
Here $z_{t+1}$ denotes a latent representation at time $t{+}1$ learned by the model, which encodes the semantic content of the next interface state. A larger Kullback Leibler (KL) divergence indicates a substantial mismatch between prediction and observation, hence the interaction provides new information and reveals a gap in the model’s internal knowledge.

To obtain a symmetric and bounded measure, we adopt the base two Jensen Shannon (JS) divergence
\begin{equation}
JS(P\parallel Q)=\tfrac{1}{2}[\sum_w P(w)\log_2\frac{P(w)}{M(w)}
+ \sum_w Q(w)\log_2\frac{Q(w)}{M(w)}]
\end{equation}
where
\begin{equation}
M(w)\;=\;\tfrac{1}{2}\big(P(w)+Q(w)\big)
\end{equation}
To better reflect the uncertainty carried by the probability mass outside the Top $K$ candidates, we introduce a tail adjustment based on the residual bucket OTHER
\begin{equation}
\begin{split}
JS^{\ast}(P\parallel Q) \;=\;&\; JS(P\parallel Q) \\
& + \lambda \cdot \tfrac{1}{2}\Big(P(\langle OTHER\rangle)+Q(\langle OTHER\rangle)\Big).
\end{split}
\end{equation}

with a tunable coefficient $\lambda>0$. The adjusted divergence accounts for the principal token distributional shift and explicitly incorporates the uncertainty from the long tail.

\begin{figure}[h]
    \centering
    \includegraphics[width=1\linewidth]{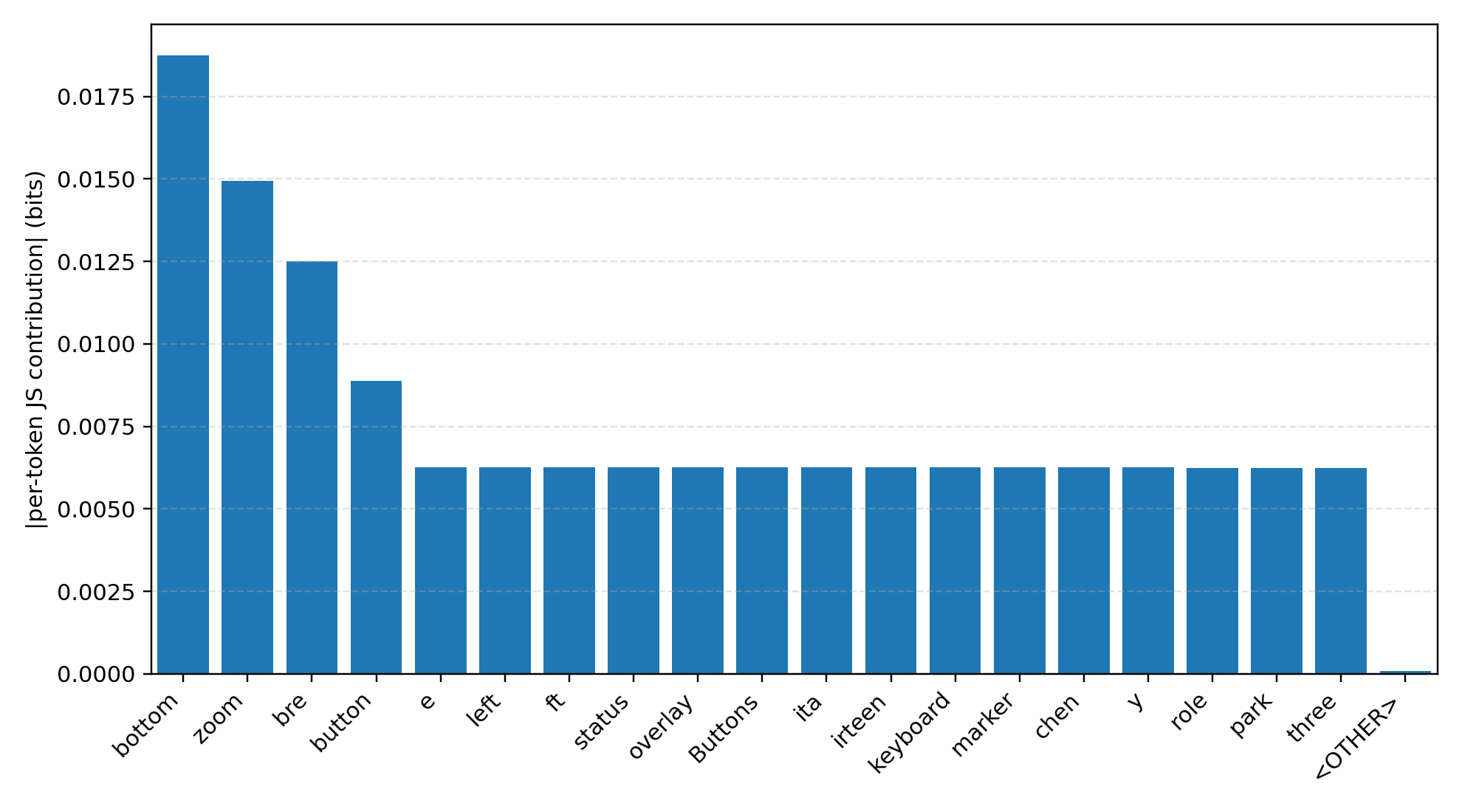}
    \caption{An example of token-level contributions to the adjusted JS divergence, showing the top 19 tokens and the residual OTHER category.}
    \label{js_contrib}
\end{figure}

We finally use the adjusted divergence as the single step information gain
\begin{equation}
I(z_{t+1};\, s_{t+1}\mid s_t,a_t)\;\approx\; JS^{\ast}(P\parallel Q)
\end{equation}
A larger value implies a greater deficiency in the model’s understanding of the application and thus a stronger curiosity for exploration. The computation pipeline is illustrated in Figure~\ref{curiosity}, which depicts the transformation from prior and posterior probabilities to the JS divergence and its tail adjusted form. In addition, Figure~\ref{js_contrib} presents an example of token-wise contributions to the divergence, highlighting which semantic elements dominate the difference between $P$ and $Q$.

\subsubsection{Cumulative Uncertainty}

To reduce the randomness of single step information gain estimates, we introduce a discounted accumulation at the application level
\begin{equation}
U(\text{app}) = \sum_{t \in T_{\text{app}}}\gamma_t \cdot JS^{\ast}(P_{t}\parallel Q_{t})
\end{equation}
where $\gamma_t$ is a weight reflecting both task difficulty and temporal decay. Once the cumulative uncertainty $U(\text{app})$ exceeds a predefined threshold $\tau$, the system proactively initiates external knowledge retrieval. Relevant information is gathered from external resources such as documentation, source code repositories, and historical trajectories, and is injected in a structured form to compensate for the model’s knowledge blind spots.

\begin{table*}[t]
\centering
\begin{threeparttable}
\caption{Comparison on \textit{AndroidWorld}. SR denotes task success rate (\%).}
\label{tab:androidworld-results}
\large
\begin{tabular}{@{}llllr@{}}
\toprule
\textbf{Method} & \textbf{Model} & \textbf{Screen Representation} & \textbf{SR} \\
\midrule
Aria-UI  \cite{yang2024aria}   & GPT-4o + Aria-UI & Screenshot                 & 44.8 \\
UI-TARS   \cite{qin2025ui}         & UI-TARS                 & Screenshot                 & 46.6 \\
AndroidGen       \cite{lai2025androidgen}         & GPT-4o                  & A11y tree                  & 46.8 \\
GUI-Explorer  \cite{xie2025gui}     & GPT-4o                  & Screenshot + A11y tree     & 47.4 \\
Agent S2            \cite{agashe2025agent}      & Agent S2                       & Screenshot                 & 54.3 \\
V-Droid      \cite{dai2025advancing}      & V-Droid (Llama-8B)       & A11y tree                  & 59.5 \\
JT-GUIAgent-V1 \cite{jtguia-v1}  & JT-GUIAgent-V1  & Screenshot                 & 60.0 \\
JT-GUIAgent-V2 \cite{JT-GUIAgent}  &   JT-GUIAgent-V2  &  Screenshot & 67.2 \\
Mobile-Agent-v3  \cite{ye2025mobile}  &   GUI-Owl-32B  &  Screenshot & 73.3 \\
MobileUse-v2     \cite{li2025mobileuse}            & Hammer-UI-32B   & Screenshot + A11y tree                & 75.0 \\
LX-GUIAgent    \cite{LX-GUIAgent2025}            & LX-GUIAgent   & Screenshot + A11y tree               & 79.3 \\
AutoGLM-Mobile \cite{xu2025mobilerl}            & AutoGLM-Mobile   & Screenshot + A11y tree                 & 80.2 \\
mobile-use \cite{li2025mobileuse}           & Llama 4-scout, Gemini 2.5 pro, GPT-5 nano   & Screenshot + A11y tree                 & 84.5 \\

\midrule
DroidRun(v0.3.3)       \cite{droidrun2025}           & Gemini-2.5-Pro           & Screenshot + A11y tree     & 63.0 \\
DroidRun(v0.3.9) & GPT-5, Gemini 2.5 Pro & Screenshot + A11y tree & 78.4\\
DroidRun(v0.3.3)+AppCards (ours) & Gemini-2.5-Pro & Screenshot + A11y tree & 69.0\\
DroidRun(v0.3.9)+AppCards (ours) & Grok-4-Fast & Screenshot + A11y tree & 60.3 \\
DroidRun(v0.3.9)+AppCards (ours) & Gemini-2.5-Pro & Screenshot + A11y tree & 71.6 \\

\textbf{DroidRun(v0.3.9)+AppCards (ours)} & \textbf{GPT-5} & \textbf{Screenshot + A11y tree} & \textbf{88.8} \\
\bottomrule
\end{tabular}
\begin{tablenotes}
\item Results are collected up to October 1, 2025, based on the \href{https://docs.google.com/spreadsheets/d/1cchzP9dlTZ3WXQTfYNhh3avxoLipqHN75v1Tb86uhHo/edit?gid=0#gid=0}{\textit{\textbf{AndroidWorld Leaderboard}}}, where our proposed method holds the \textbf{top position}
\end{tablenotes}
\end{threeparttable}
\end{table*}

\subsection{Retrieving and Embedding External Knowledge}

When the cumulative uncertainty $U(\text{app})$ exceeds the predefined threshold $\tau$, the agent proactively triggers external knowledge retrieval in order to compensate for its insufficient understanding of application functionalities. Unlike approaches that rely solely on internal parametric knowledge, this mechanism explicitly incorporates external resources into the reasoning loop, thereby providing additional structured support in high uncertainty scenarios.

We consider three heterogeneous sources of knowledge. The first is online documentation $D_{\text{web}}$, which includes developer manuals, API documentation, and user guides. The second is source code and version history $D_{\text{git}}$, which covers function definitions, dependency structures, and development logs. The third is historical interaction trajectories $D_{\text{traj}}$, which record execution paths and experiential traces. Once retrieval is triggered, the system uses the uncertainty signal $U(\text{app})$ together with the threshold $\tau$ to form targeted queries, and consolidates candidate knowledge units from these multimodal sources
\begin{equation}
\text{Card}(App)= \text{Retrieve}(D_{\text{web}} \cup D_{\text{git}} \cup D_{\text{traj}}, U(\text{app}), \tau)
\end{equation}

To efficiently integrate the retrieved knowledge, we employ a structured representation referred to as AppCards. Each AppCard is constructed for a single application with the objective of compressing unstructured information into modular knowledge units that are convenient for reasoning. A typical AppCard consists of four components. The first component is functional semantics, which describes the purpose and applicable scope of API calls. The second is input and output constraints, which specify parameter types, return values, and boundary conditions. The third is the mapping between interface elements and functionalities, which aligns user interface components such as button texts with their triggered functions. The fourth is a set of common interaction patterns and exception handling strategies. Through a unified template design, AppCards achieve multi-source coverage while ensuring interpretability and reusability of the knowledge representation.

During task execution, the agent first identifies which applications need to be invoked, then retrieves the corresponding AppCards and injects them into the system prompt. In this manner, uncertainty is operationalized as a curiosity signal, with information gain measured through the divergence between prior and posterior distributions. Once cumulative uncertainty exceeds the threshold, the system activates external retrieval and incorporates the structured AppCard representations into the reasoning process, thereby mitigating knowledge deficiencies and enhancing the agent’s generalization capability.

\begin{table*}[h]
\centering
\large
\caption{Performance comparison of DroidRun and DroidRun+AppCards across difficulty levels. The numbers indicate successes out of total tasks, with success rates and relative improvements.}
\label{tab:ablation}

\begin{tabular}{l c | c c | c c | c c}
\hline
\textbf{Model} & \textbf{Method} & 
\multicolumn{2}{c|}{\textbf{Easy (61)}} & 
\multicolumn{2}{c|}{\textbf{Medium (36)}} & 
\multicolumn{2}{c}{\textbf{Hard (19)}} \\
 & & \textbf{Succ.} & \textbf{Rate} & \textbf{Succ.} & \textbf{Rate} & \textbf{Succ.} & \textbf{Rate} \\
\hline
Gemini-2.5-Pro & DroidRun(v0.3.3) & 48 & 78.7\% & 20 & 55.6\% & 5 & 26.3\% \\
Gemini-2.5-Pro & DroidRun(v0.3.3)+AppCards & 51 & 83.6\% ($\uparrow$4.9\%) & 22 & 61.1\% ($\uparrow$5.5\%) & 7 & 36.8\% ($\uparrow$10.5\%) \\
\hline
Grok-4-fast & DroidRun(v0.3.9) & 43 & 70.5\% & 18 & 50.0\% & 10 & 52.6\% \\
Grok-4-fast & DroidRun(v0.3.9)+AppCards & 44 & 72.1\% ($\uparrow$1.6\%) & 20 & 55.6\% ($\uparrow$5.6\%) & 6 & 31.6\% ($\downarrow$21.0\%) \\
\hline
Gemini-2.5-Pro & DroidRun(v0.3.9) & 48 & 78.7\% & 19 & 52.8\% & 9 & 47.4\% \\
Gemini-2.5-Pro & DroidRun(v0.3.9)+AppCards & 51 & 83.6\% ($\uparrow$4.9\%) & 22 & 61.1\% ($\uparrow$8.3\%) & 10 & 52.6\% ($\uparrow$5.2\%) \\
\hline
GPT-5 & DroidRun(v0.3.9) & 55 & 90.2\% & 32 & 88.9\% & 11 & 57.9\% \\
GPT-5 & DroidRun(v0.3.9)+AppCards & 56 & 91.8\% ($\uparrow$1.6\%) & 32 & 88.9\% ($\uparrow$0.0\%) & 15 & 78.9\% ($\uparrow$21\%) \\
\hline
\end{tabular}

\end{table*}

\section{Experimental}
\subsection{Experimental Settings}

\textbf{Benchmark.}  
Experiments are conducted on the \textit{AndroidWorld} benchmark, which includes 116 interactive tasks across about twenty widely used Android applications. Tasks are instantiated from parameterized templates to evaluate both proficiency and generalization. The benchmark spans scheduling, configuration, note taking, finance, messaging, media, and web navigation. Tasks are grouped into three difficulty levels: 61 easy tasks with short single-app interactions, 36 medium tasks involving multi-step creation or editing, and 19 hard tasks requiring extended reasoning or cross-app coordination. We report task success rate, defined as completing the goal within a fixed interaction budget. Unless otherwise specified, agents observe both screenshots and accessibility trees at each step.  

\textbf{Implementation details.}
All evaluations use the open-source \textit{DroidRun} framework for stable execution on \textit{AndroidWorld}. We consider two releases (v0.3.3 and v0.3.9) to assess robustness to infrastructure changes. Unless otherwise specified, Gemini 2.5 Pro is the default backbone (temperature 0.3); we also report results with GPT 5 and Grok 4 fast under v0.3.9. Our system augments DroidRun with a curiosity-driven knowledge retrieval module: a curiosity score estimates uncertainty about functional knowledge and triggers external retrieval when it exceeds a fixed threshold. Retrieved information from application documentation and source repositories is consolidated into structured \textit{AppCards} and selectively incorporated into reasoning and action planning. All methods use the same step budget, observation modality, and environment configuration for fair comparison.

\begin{figure*}[h]
    \centering
    \includegraphics[width=1\linewidth]{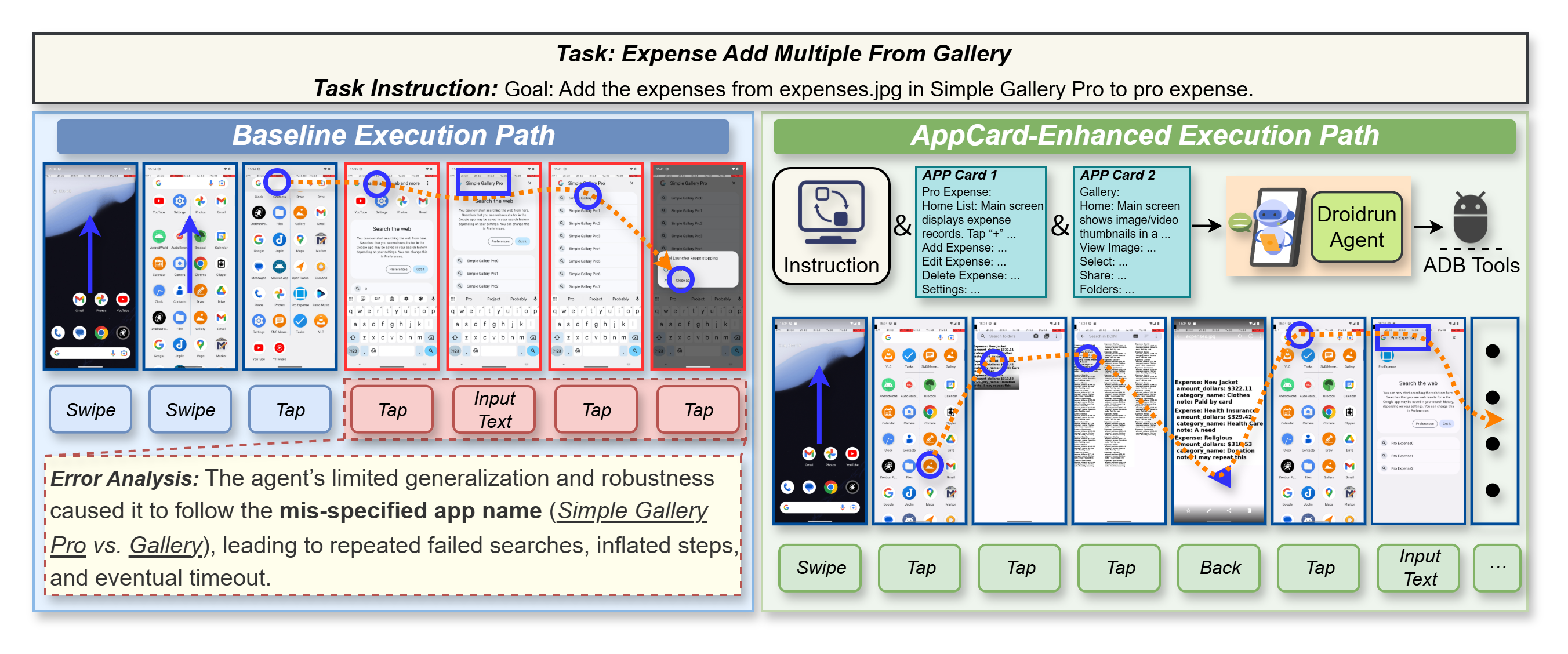}
    \caption{Case study of the task Expense Add Multiple From Gallery. The baseline path on the left fails due to application name ambiguity, while the AppCard enhanced path on the right leverages structured knowledge to enable stable and successful task execution.}
    \label{casestudy}
\end{figure*}

\subsection{Results}

Table~\ref{tab:androidworld-results} presents the overall comparison on the AndroidWorld benchmark. Compared with existing state-of-the-art methods, DroidRun with AppCards using GPT 5 in version v0.3.9 achieves a success rate of \textbf{88.8}\%, surpassing the previously reported \textbf{84.5}\% of mobile use and establishing a new publicly available best result. In terms of relative improvement, GPT 5 increases from \textbf{84.5}\% to \textbf{88.8}\%, corresponding to an absolute gain of \textbf{4.3} percentage points and a relative increase of approximately \textbf{5.1}\%. For Gemini 2.5 Pro, the performance in version v0.3.3 improves from \textbf{63.0}\% to \textbf{69.0}\%, and in version v0.3.9 improves from \textbf{65.5}\% to \textbf{71.6}\%. Both versions yield gains of around \textbf{6} percentage points, with relative improvements in the range of \textbf{9 to 10}\%. In contrast, Grok 4 fast decreases from \textbf{61.2}\% to \textbf{60.3}\% in version v0.3.9, showing no benefit from AppCards.

Taken together, these results indicate that the effectiveness of AppCards varies across models and versions. When the backbone model has strong reasoning and knowledge integration capabilities, AppCards consistently translate into performance improvements, with GPT 5 achieving breakthrough results. For weaker or stylistically mismatched models, however, AppCards may not be fully utilized and can even introduce negative effects. This suggests that AppCards not only contribute to performance enhancement but also expose critical interactions between external knowledge and model characteristics, offering insights for designing more robust knowledge injection mechanisms.

\subsection{Ablation Study}

Table~\ref{tab:ablation} reports detailed performance across different task difficulty levels. In version v0.3.3, Gemini 2.5 Pro combined with AppCards improves from \textbf{78.7}\% to \textbf{83.6}\% on easy tasks, from \textbf{55.6}\% to \textbf{61.1}\% on medium tasks, and from \textbf{26.3}\% to \textbf{36.8}\% on hard tasks. Although improvements appear across all levels, the most substantial gain is observed in hard tasks, with an increase of \textbf{10.5} percentage points, highlighting the value of AppCards in guiding models through complex task structures. In version v0.3.9, GPT 5 exhibits even more pronounced improvements. Easy tasks remain nearly unchanged, moving from \textbf{90.2}\% to \textbf{91.8}\%. Medium tasks remain stable at \textbf{88.9}\%. Hard tasks, however, increase dramatically from \textbf{57.9}\% to \textbf{78.9}\%, yielding a \textbf{21.0} percentage point improvement. This underscores the critical role of AppCards in bridging knowledge gaps under complex conditions. Gemini 2.5 Pro in the same version shows consistent improvements of \textbf{4.9}, \textbf{8.3}, and \textbf{5.2} percentage points across easy, medium, and hard tasks, respectively, reflecting balanced benefits. By contrast, Grok 4 fast shows modest gains on easy and medium tasks but experiences a sharp decline on hard tasks, dropping from \textbf{52.6}\% to \textbf{31.6}\%, a decrease of \textbf{21.0} percentage points, indicating that its knowledge integration ability is insufficient to leverage AppCards in demanding scenarios.

Overall, these stratified experiments demonstrate that AppCards provide the greatest benefit in complex tasks, substantially enhancing long-horizon reasoning and cross-application operations. At the same time, the experiments reveal clear divergences among models. Strong backbones are able to reliably absorb external knowledge and achieve significant gains, whereas weaker models may encounter instability. This contrast shows that AppCards not only improve average performance but also provide concrete evidence for understanding the mechanisms of interaction between external knowledge and model capabilities.


\section{Case Study}

\textbf{Task and Challenge: }  
The examined task requires the agent to transfer the file \textit{expenses.jpg} from \textit{Simple Gallery Pro} into \textit{Pro Expense}. This operation involves coordinated interactions across multiple applications and therefore places high demands on the model’s ability to disambiguate entities, identify functional entry points, and reason over inter-application paths. Baseline experiments reveal that in the absence of external knowledge the model is unable to complete the task reliably. The agent frequently confuses the system gallery with the intended application, repeatedly searches within the app drawer, and generates redundant or erroneous actions. These behaviors accumulate until the interaction budget is exhausted, highlighting fundamental deficiencies in semantic grounding and strategic planning across heterogeneous applications.

\textbf{AppCard Injection and Path Optimization: }  
AppCards are designed to inject structured functional semantics into the agent in order to mitigate uncertainty in the decision process. For this case, the Pro Expense home screen is represented with explicit entries for record creation and editing, while the Simple Gallery Pro interface is summarized with functional elements such as thumbnail browsing, image selection, and sharing options. This knowledge is consolidated into a compressed frame that provides the agent with direct mappings between observed states and feasible actions. As a result, the decision sequence is no longer driven by exploratory search but evolves into a stable and logically ordered trajectory. The redundant and error-prone behaviors observed in the baseline are thus eliminated, and the agent follows a path that is both shorter and more predictable.

\textbf{Analysis: }  
The case study demonstrates how AppCards provide structured knowledge that directly addresses the limitations of baseline agents. By encoding the correct application identity and presenting explicit functional entry points, AppCards resolve entity-level ambiguity and stabilize application selection. By abstracting key visual elements, such as thumbnails, share icons, and form fields, into operational anchors, they transform interface features into consistent functional cues that guide execution. In parallel, AppCards constrain the decision space by reducing an open-ended search into a set of contextually appropriate candidates, aligning visual recognition with functional reasoning at critical decision points. This combined support allows the agent to transition from fragmented trial and error toward coherent plan execution. The outcome is reflected in shorter trajectories, improved stability in planning, and reliable completion of the cross-application transfer. As illustrated in Figure~\ref{casestudy}, the contrast between the baseline trajectory and the AppCard-enhanced path highlights the role of structured knowledge in enabling robust and efficient task completion.

\section{Discussion}
The curiosity mechanism provides a principled means to estimate and quantify an agent’s insufficient understanding during task execution. By formalizing epistemic uncertainty as a numerical score, it allows the differentiation between failures caused by missing functional knowledge and those resulting from alternative factors such as interface variability or execution errors. This diagnostic role supports more accurate interpretation of outcomes and prevents the misattribution of errors to model limitations alone.

In addition to diagnosis, curiosity also drives the automated construction of AppCards. When uncertainty exceeds a threshold, it identifies specific knowledge gaps and triggers targeted retrieval from external sources. The retrieved information is consolidated into structured AppCards that directly address the deficiencies observed during execution. In this way, curiosity transforms external knowledge retrieval into a selective and context-sensitive process, simultaneously explaining failure and guiding the generation of knowledge resources that enhance future performance.

\section{Conclusion}
This paper presents a curiosity driven knowledge retrieval and augmentation framework for mobile agents in complex applications. Curiosity quantifies epistemic uncertainty during execution and triggers targeted retrieval, while AppCards consolidate external information into structured functional knowledge that can be injected into reasoning and planning. Evaluations on AndroidWorld under a fixed interaction budget and observation modality show consistent gains across backbone models, reaching a new state of the art success rate of 88.8\% when combined with GPT 5. The improvements are most pronounced on tasks that require multi step reasoning and cross application coordination, where planning becomes more stable and trajectories are shorter. Limitations include model dependence and the need for calibrated triggering and snippet selection. Future work will explore adaptive thresholding and granularity control, alignment strategies for different backbones, automated construction and versioning of AppCards, and cost aware retrieval scheduling. Overall, the framework provides a practical path toward knowledge aware and robust mobile agents.

\section*{Acknowledgments}
This work is supported in part by Shanghai Municipal Natural Science Foundation (23ZR1425400) and Shanghai Soft Science Project (25692114700).


\bibliographystyle{ACM-Reference-Format}
\balance
\bibliography{refs}

\appendix

\begin{figure*}[h]
    \centering
    \includegraphics[width=\linewidth]{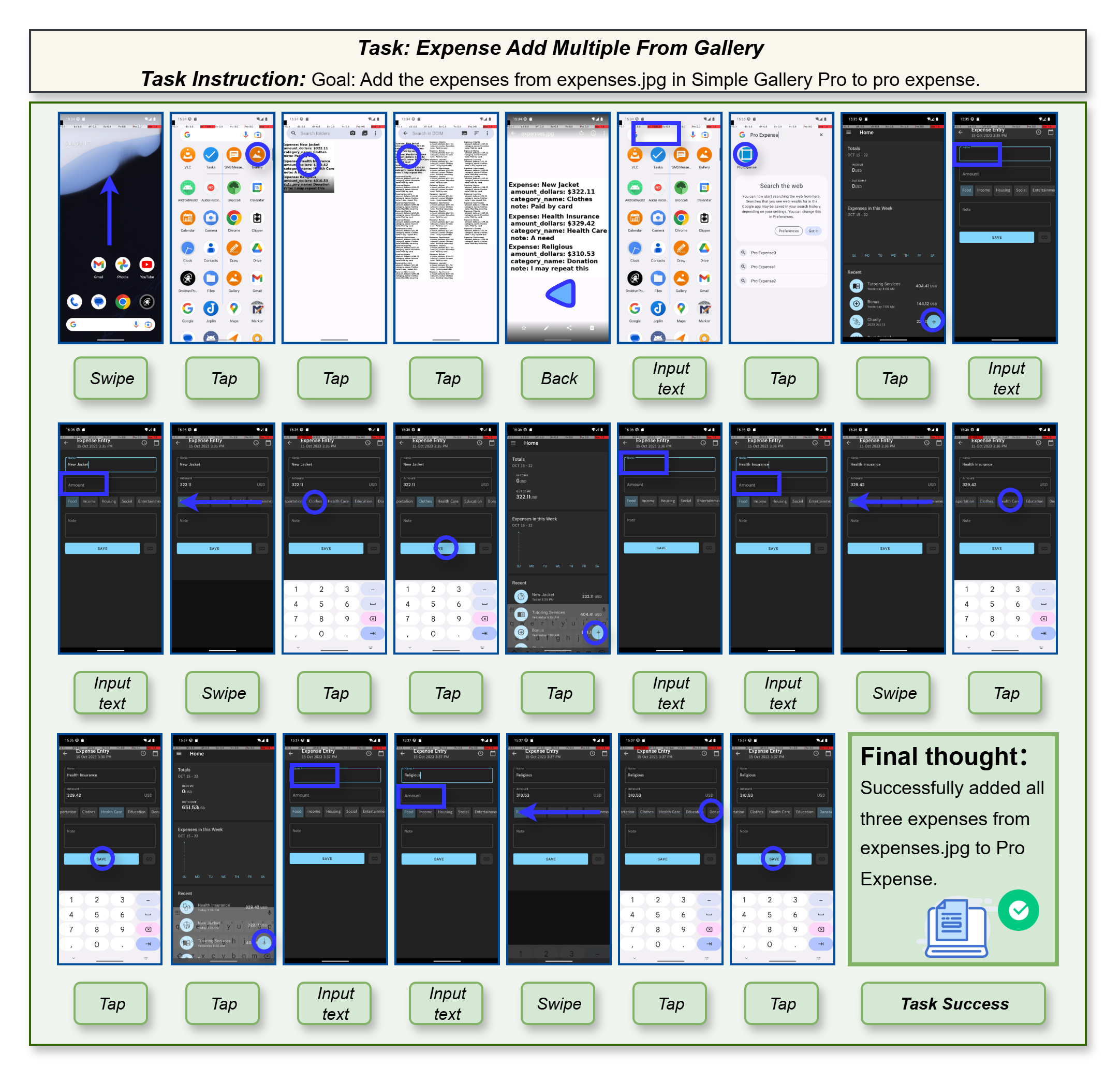}
    \caption{Trajectory example for the \textit{Expense Add Multiple From Gallery} task. The agent completes the task by importing all expenses from the gallery image into the Pro Expense application.}
    \label{fig:traj-expense}
\end{figure*}

\begin{figure*}[ht]
    \centering
    \includegraphics[width=\linewidth]{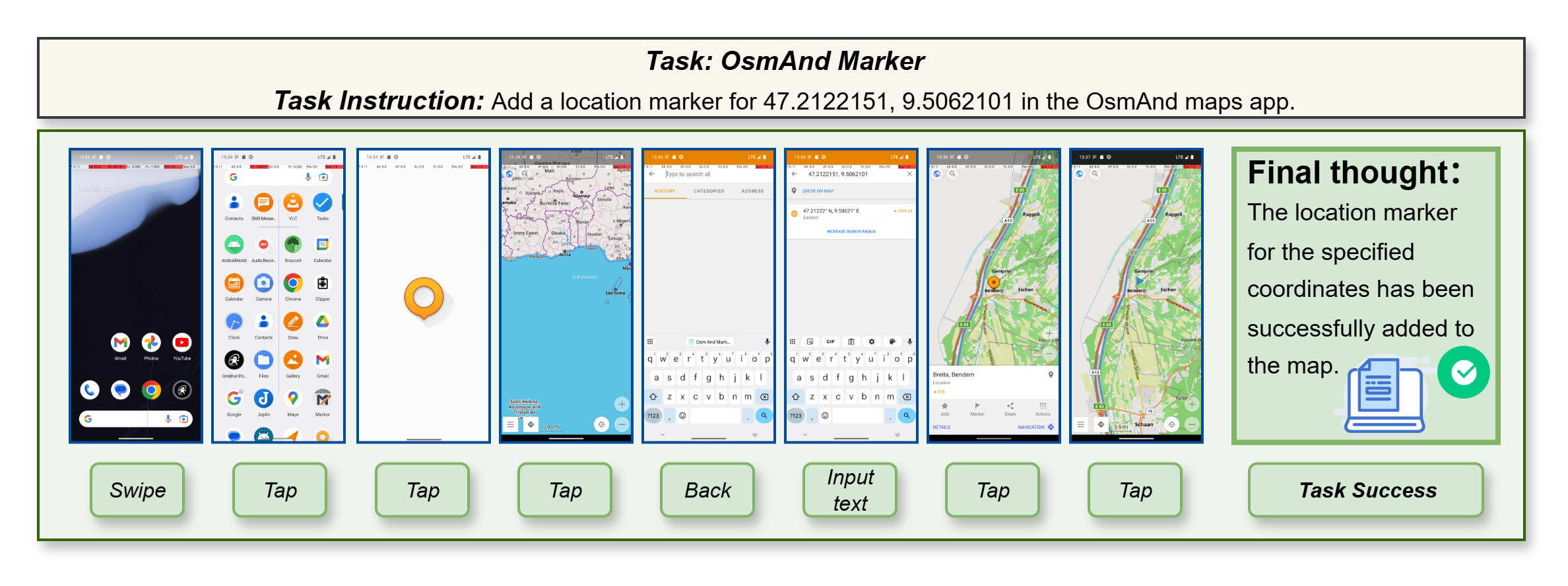}
    \caption{Trajectory example for the \textit{OsmAnd Marker} task. The agent successfully adds the location marker for the specified coordinates.}
    \label{fig:traj-osm}
\end{figure*}

\section{Trajectory Example}
\label{app:trajectory-example}

To provide a concrete illustration of how our framework executes mobile tasks, we present the complete trajectories of two representative examples that also appear in the main paper. Each trajectory records the full sequence of interface states and actions, together with the final verification outcome.

\subsection*{Expense Add Multiple From Gallery Task}
Figure~\ref{fig:traj-expense} presents the trajectory for the task \textit{Expense Add Multiple From Gallery}. The instruction requires adding all expense records from an image file (\texttt{expenses.jpg}) stored in the gallery into the Pro Expense application. The trajectory demonstrates how the agent switches between applications, selects the correct image, and systematically inputs the extracted expenses. The final verification shows that all three expenses are correctly recorded in Pro Expense, confirming successful task completion.

\subsection*{OsmAnd Marker Task}
Figure~\ref{fig:traj-osm} shows the trajectory for the task \textit{OsmAnd Marker}, where the instruction is to add a location marker for the specified coordinates in the OsmAnd maps application. The trajectory begins with the agent launching the application, navigating through the interface, entering the coordinates, and confirming the marker placement. The final state indicates successful task completion, with the marker correctly added on the map.

\section{AppCard Specifications}
\label{appendix:appcards}

In this appendix we provide representative \textit{AppCards} that were constructed for several applications used in our experiments. Each AppCard encodes the key functionalities of an application into structured modules, serving as reusable building blocks for mobile agent reasoning. The following cases illustrate how task-relevant knowledge is formalized.  

\subsection*{Gallery}
The \textit{Gallery} application provides image and video management. Its AppCard focuses on viewing, selecting, sharing, and folder navigation.

\begin{lstlisting}[language=,basicstyle=\ttfamily\small,frame=single,breaklines=true]
### Gallery:
1. Home: Main screen shows image/video thumbnails in grid; tap any thumbnail to open full view.
2. ViewImage: In full view, swipe left/right to browse; pinch to zoom; toolbar has Share, Edit, Delete.
3. Select: Long-press a thumbnail to select one or multiple images; then actions (Share, Delete, Info) appear in top bar.
4. Share: After selecting image(s), tap 'Share' icon to send to another app.
5. Folders: Accessible via side menu or top bar; lets you switch between Albums/Folders containing media.
\end{lstlisting}

\subsection*{VLC}
The \textit{VLC} application is a versatile media player supporting local and network playback, streaming, and extensive customization. Its AppCard highlights playback control, media management, and advanced features.

\begin{lstlisting}[language=,basicstyle=\ttfamily\small,frame=single,breaklines=true]
### VLC:
1. Open/Play: Open local files, discs, devices, and network streams; plays almost anything.
2. Playback Controls: Play/Pause, seek/jump, next/previous; on-screen and hotkey controls.
3. Subtitles & Audio Tracks: Load/select external subtitle files; switch audio tracks; adjust sync.
4. Playlist: Queue and manage multiple items; reorder and continuous play.
5. Streaming/Convert: Convert/transcode media and stream output to various formats/protocols.
6. Video/Audio Effects: Apply video filters (deinterlace, rotate, crop, etc.) and audio equalizer.
7. Network & Casting: Open network URLs; Chromecast streaming (version 3.0+).
8. High-Res Playback: Hardware-accelerated decoding; up to 4K/8K, 10-bit, HDR (platform-dependent).
9. Snapshots & Frame Step: Capture still frames (snapshots) and step through frames for inspection.
10. Preferences: Extensive settings for codecs, subtitles, audio/video output, hotkeys, and interfaces.
\end{lstlisting}

\subsection*{OsmAnd}
The \textit{OsmAnd} application is a navigation and offline maps tool. Its AppCard encodes map viewing, routing, track recording, and plugin extensions.

\begin{lstlisting}[language=,basicstyle=\ttfamily\small,frame=single,breaklines=true]
### OsmAnd:
1. **Map Viewing / Offline Maps:** Download maps for regions; view maps offline without internet; support different map styles (topographic, contour, shading).  
2. **Search & POI:** Search places by address/name/coordinates; browse POIs (restaurants, ATMs, etc.) and categories.  
3. **Navigation / Routing:** Plan routes for car, bicycle, pedestrian; support turn-by-turn guidance; voice prompts; automatic rerouting if deviated.  
4. **Track Recording:** Record your trip/track (GPX format); view recorded paths on the map.  
5. **Plugins / Extensions:** Use plugin features (e.g. contour lines, hill shading, Mapillary, public transport stops) to extend map display or routing options.  
6. **Offline Functionality:** All key features (map view, routing, search) work offline once maps are downloaded.  
7. **Settings / Profiles:** Configure vehicle profile (car/bike/pedestrian), map updates, display settings, map styles, routing preferences.  
8. **Map Interactions:** Zoom, pan, rotate map; set waypoints or intermediate stops; save favorite places.
\end{lstlisting}

\subsection*{Broccoli}
The \textit{Broccoli} application is a recipe management tool designed for browsing, adding, importing, and assisting in cooking workflows. Its AppCard emphasizes recipe organization, seasonal ingredient guidance, and offline accessibility.

\begin{lstlisting}[language=,basicstyle=\ttfamily\small,frame=single,breaklines=true]
### Broccoli:
1. Recipe List / Library: Browse your recipe collection; view recipes by category/hashtag; search by title or keywords.
2. Add Recipe: Create a new recipe with fields: title, ingredients, instructions/steps; optional image; set category/tags.
3. Import Recipe: Import recipes from blogs/websites when recipe metadata (JSON-LD) is available; also import from backups.
4. Cooking Assistant: Use fullscreen mode to follow steps; adjust ingredient quantities; distraction-free cooking view.
5. Seasonal Calendar: View seasonal ingredients for your region; highlight seasonal items in recipes; search by season.
6. Backup & Restore / Offline Access: Save recipe backups; restore; access all saved recipes without internet.
7. Settings & Preferences: Configure theme (e.g. dark mode), units, recipe sorting, display preferences.
8. Permissions & Storage: Request storage read/write; manage images/photos; ensure data stored on device.
\end{lstlisting}

\subsection*{Contacts}
The \textit{Contacts} application provides management of personal and professional information through a structured contact list, detail views, and customizable settings. Its AppCard highlights the main operations including creation, modification, deletion, and search, together with account preferences and permission management that ensure consistent and reliable usage across devices.  

\begin{lstlisting}[language=,basicstyle=\ttfamily\small,frame=single,breaklines=true]
### Contacts:
1. Home / Contact List: Displays all contacts; scroll through alphabetical list; tap a contact for details.  
2. Search: Use search bar or search icon from Home to find contacts by name or phone number.  
3. View Contact Details: In contact detail screen, see name, phone numbers, email addresses, photo, other fields.  
4. Add Contact: Tap "+" or "Add" -> input name, phone number, optionally email/address/photo -> save.  
5. Edit Contact: In detail view, tap edit icon -> change fields (name, phone, email, etc.) -> save changes.  
6. Delete Contact: From the detail screen or list (long-press) choose "Delete" -> confirm.  
7. Permissions: Request required permissions (Contacts, Storage if needed for photos).  
8. Settings / Default Account: Choose default account for new contacts (if device supports multiple accounts); set display/order preferences.  
\end{lstlisting}

\subsection*{Pro Expense}
The \textit{Pro Expense} application manages personal expense tracking. The AppCard captures core modules ranging from adding and editing expenses to statistics and backup functionality.  

\begin{lstlisting}[language=,basicstyle=\ttfamily\small,frame=single,breaklines=true]
### Pro Expense:
1. **Home/Expenses:** Main list of expense items; tap to view or edit; long-press for actions (Delete, Duplicate).
2. **AddExpense (FAB +):** Create a new expense with fields such as **Amount**, **Category**, **Date/Time**, and **Note**; save to return to the list.
3. **Edit/Delete:** Open an expense -> tap **✎** to edit; use menu/options to **Delete**.
4. **Categories:** Choose a category when adding/editing an expense; categories power the stats views.
5. **Stats:** View weekly and category statistics (simple graphs/summary) from the stats section.
6. **Search/Filter:** From the list, search by text or filter by date/category to locate expenses.
7. **Share-Into (Import Text):** Receive plain text via Android **Share** from other apps (e.g., a note app) to prefill/parse expense content, then confirm to add items.
8. **Backup/Restore (Local):** Manage on-device backups/restore (cloud sync not supported).
9. **Settings:** Adjust language, theme/appearance, and general app preferences (privacy, performance).
\end{lstlisting}

\subsection*{Markor}
The \textit{Markor} application is a Markdown-based note-taking and text-editing app. The AppCard highlights its creation, editing, search, and sharing features.  

\begin{lstlisting}[language=,basicstyle=\ttfamily\small,frame=single,breaklines=true]
### Markor:
1. Home: List of notes/documents; tap to open folder or file.
2. CreateNote: Bottom-right '+' -> new note -> editor -> Save.
3. EditNote: Tap existing note -> modify in Markdown editor -> Save.
4. DeleteNote: Long-press note -> Delete -> confirm.
5. Search: Tap search icon -> enter keyword -> view matches.
6. Settings: Adjust theme, syntax highlighting, font, default folder.
7. Share: Open note -> share icon -> choose target app.
\end{lstlisting}

Together, these AppCards demonstrate how heterogeneous applications can be abstracted into a unified representation that supports modular reasoning and reliable task execution by mobile agents.

\end{document}